\documentclass[10pt,twocolumn,letterpaper]{article}

\usepackage[pagenumbers]{wacv} 

\usepackage[utf8]{inputenc}
\usepackage[T1]{fontenc}
\usepackage{url}
\usepackage{booktabs}   
\usepackage{colortbl}   
\usepackage[dvipsnames,table]{xcolor}  
\usepackage{multirow}   
\usepackage{amsfonts}
\usepackage{amsmath}
\usepackage{amssymb}
\usepackage{graphicx}
\usepackage{microtype}

\definecolor{wacvblue}{rgb}{0.21,0.49,0.74}
\usepackage[breaklinks,colorlinks,allcolors=wacvblue]{hyperref}

\title{LeFlow: Generative Latent Flow Planning for World Models}

\author{
Hsiang-Wei Huang,
Jianxu Shangguan,
Junbin Lu,
Jenq-Neng Hwang \\
University of Washington, United States \\
\texttt{\{hwhuang,jxb1st,junbinlu,hwang\}@uw.edu}
}

\begin{document}
\maketitle

\begin{abstract}
Latent world models are inherently strong encoders that transform image pixel to latent embedding,
yet existing world models still rely on online trajectory
optimization for action planning: for every state--goal pair, an iterative optimizer is run from scratch to search for optimal action sequences, treating the world model as a
black-box simulator. This approach pays the full iterative optimization cost
anew at every replanning step and reuses no planning experience across queries.
In this work, we ask whether planning itself can be amortized once a latent
world model has been learned. We present LeFlow, which learns a
reusable latent trajectory prior operating directly in the latent dynamics
space from the world model. LeFlow recasts planning as conditional
latent trajectory generation: a rectified-flow model imagines a future latent
path between the current and goal embeddings, an inverse dynamics decoder turns
latent transitions into action chunks, and the frozen world model verifies each
candidate by autoregressive rollout. Across four major goal-conditioned
pixel-control benchmarks, LeFlow replaces iterative action-space optimization
with amortized latent planning and fixed-budget rollout selection, achieving
consistent success-rate gains with an order-of-magnitude reduction
in planning time. Our results argue that
latent world models should support not only prediction but reusable planning
priors. Our code is available at
\url{https://github.com/hsiangwei0903/LeFlow}.

\end{abstract}


\section{Introduction}
\label{sec:intro}

World models let agents plan by simulating the consequences of their actions
before acting~\citep{ha2018worldmodels}. Joint-Embedding Predictive
Architectures (JEPAs)~\citep{lecun2022path} make this practical by modeling
dynamics in a compact latent space: observations are encoded into
low-dimensional latents and a predictor rolls the latent state forward
conditioned on actions. LeWorldModel (LeWM)~\citep{maes2026lewm} is a recent
JEPA that trains stably end-to-end from raw pixels with a single regularization
hyperparameter and is competitive across diverse 2D and 3D control tasks. Latent
world models, in short, have become strong \emph{predictors}.

\begin{figure}[tbp]
  \centering
  \includegraphics[width=\linewidth]{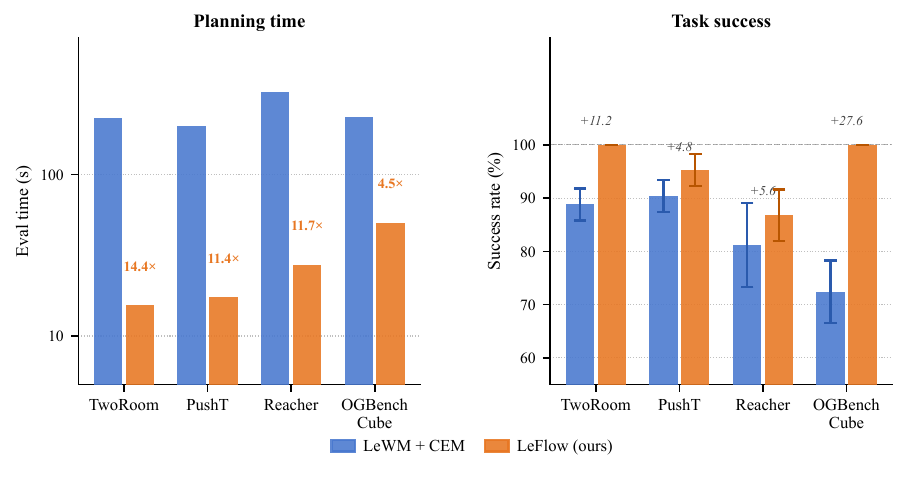}
  \caption{\textbf{LeFlow plans an order of magnitude faster than CEM at
  better success.} End-to-end evaluation time (left) and success rate
  (right) for LeWM+CEM and LeFlow under the same protocol ($H{=}5$, $50$
  episodes). Replacing online black-box search with an amortized latent
  trajectory prior removes the per-step optimization loop while preserving task
  performance.}
  \label{fig:runtime}
\end{figure}

Planning on top of them, however, has not advanced at the same pace. Even with a
well-trained latent world model in hand, planning is still posed as online
trajectory optimization: for every state--goal pair, an optimizer searches for a
raw action sequence whose autoregressive latent rollout terminates near the
encoded goal. LeWM, for instance, solves this finite-horizon optimal-control
problem at inference with the Cross-Entropy Method
(CEM)~\citep{rubinstein1999cem}. CEM is a practical optimizer for
low-dimensional, smooth objectives. Our focus is its repeated online cost: CEM
treats the world model as a black-box simulator and solves every state--goal
query \emph{from scratch}: the world model is queried thousands of times per
replanning step yet never contributes structural knowledge about which
trajectories are worth considering, and no effort spent planning one query is
reused for the next.

We propose to amortize planning itself. Rather than repeatedly solving an
optimization problem online, we learn -- once, offline -- a reusable
\emph{latent trajectory prior} directly in the frozen world model's latent
space. A frozen world model already contains the structure needed: its encoder
organizes observations geometrically and its predictor defines which latent
transitions are dynamically reachable. Planning should exploit that structure,
not rediscover it through black-box search.
We recast planning as \emph{conditional latent trajectory generation}: rather
than searching for actions, we generate a latent path and decode the actions that
realize it afterwards. Action trajectories are low-level and highly multimodal --
many action sequences realize nearly the same state trajectory -- whereas latent
state trajectories are smoother and geometrically structured, so the reusable
planning structure lives in how the latent state should evolve.

We instantiate this in \textbf{LeFlow}, a lightweight planner on top of a
\emph{frozen} LeWM. LeFlow has three modular components: (1) a
\emph{rectified-flow latent planner}~\citep{liu2023rectified,lipman2023flow}
that generates the interior of a latent path conditioned on the start and goal
embeddings -- the amortized prior; (2) an \emph{inverse dynamics} decoder that
converts each latent transition into an executable action chunk, separating
planning from control; and (3) a \emph{rollout verification} stage that scores
each candidate by its \emph{actual} frozen-LeWM rollout distance to the goal
rather than the generated (clamped) endpoint, projecting generative proposals
back onto the manifold of trajectories the world model can actually control. On LeWM, it yields an order-of-magnitude
reduction in planning time at better success across four benchmarks, as shown
in Figure~\ref{fig:runtime}. Our contributions are:
\begin{itemize}
  \item We recast latent planning as conditional latent trajectory generation,
        and show that planning knowledge can be \emph{amortized} -- learned
        once from offline trajectories into a reusable latent trajectory prior,
        rather than re-solved online for every state--goal pair.
  \item We show that latent world models contain sufficient structure for learned planning: LeFlow combines generative latent trajectory proposals with rollout-based verification \emph{inside} the
        original world model, without modifying it.
  \item Across four benchmarks, amortized latent
        planning replaces iterative action-space optimization with fixed-budget
        rollout selection while \emph{improving} success, cutting end-to-end
        planning time by roughly an order of
        magnitude and generalizing to held-out episodes; ablations show that
        latent-path generation beats direct action generation and that rollout
        verification is essential for dynamically feasible plans.
\end{itemize}

\section{Related Work}
\label{sec:related}

\paragraph{Latent world models.}
Latent world models learn dynamics in a compressed embedding space and plan by
rolling out simulated futures~\citep{ha2018worldmodels,hafner2020dreamer,hafner2021dreamer2}.
Test-time planning typically uses CEM~\citep{rubinstein1999cem,pinneri2021icem} or
MPPI~\citep{williams2017mppi} as black-box optimizers over the world model.
JEPA-style models~\citep{lecun2022path} predict in latent space without pixel
reconstruction; this family includes image-level I-JEPA~\citep{assran2023ijepa}
and video-level V-JEPA~\citep{bardes2024vjepa}, which demonstrate that
joint-embedding prediction yields rich visual representations without
generative decoding. Extending JEPAs to control requires additionally modeling
action-conditioned dynamics; prior world-model variants rely on pretrained
encoders~\citep{zhou2024dinowm} or heuristic multi-term
objectives~\citep{sobal2025pldm}. LeWM~\citep{maes2026lewm} addresses both and
is our frozen backbone. We ask whether the learned latent structure already
encodes what action planning recovers -- making iterative optimization unnecessary.

\begin{figure*}[tbp]
  \centering
  \includegraphics[width=\linewidth]{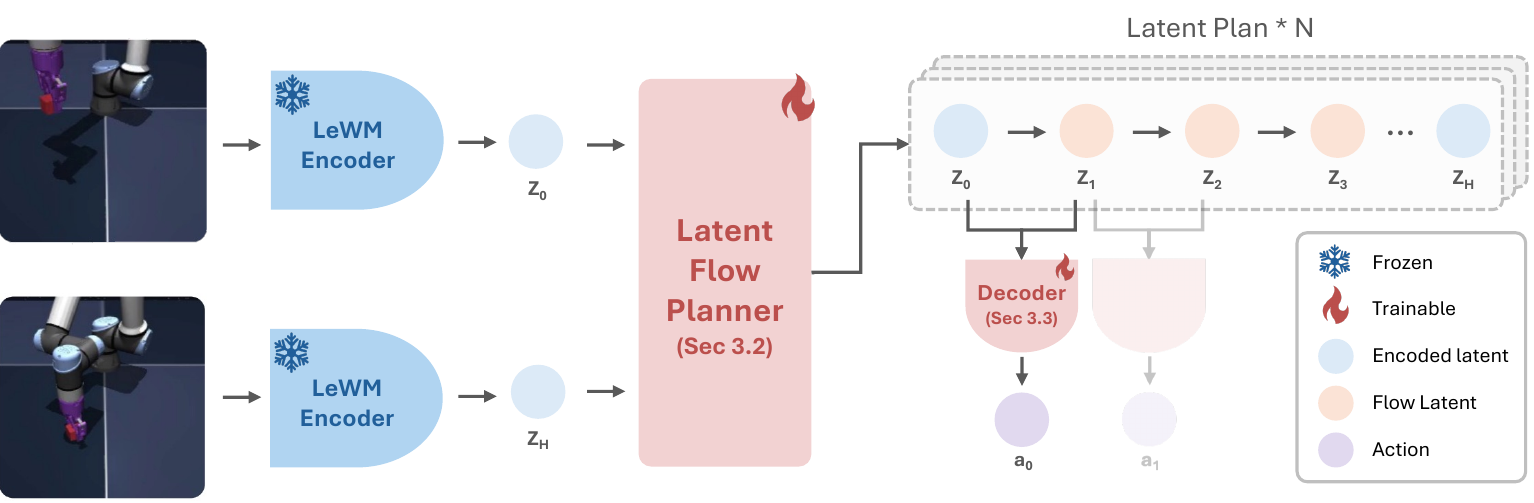}
  \caption{\textbf{LeFlow architecture.} A frozen LeWM encoder maps the current
  and goal observations to latent anchors $z_{\text{start}}$ and
  $z_{\text{goal}}$. A conditional rectified-flow planner generates the interior
  of a latent path between them; an inverse dynamics decoder converts each
  latent transition into an action chunk. Only the flow
  planner and inverse dynamics decoder are trained; LeWM encoder is frozen throughout.}
  \label{fig:method}
\end{figure*}

\paragraph{Amortized planning and inverse dynamics.}
Inverse-dynamics models -- predicting the action responsible for a latent
transition -- appear in model-based RL and representation
learning~\citep{agrawal2016poking,pathak2017curiosity,baker2022vpt}.
GLAMOR~\citep{paster2021glamor} is a notable precursor that learns a
goal-conditioned recurrent inverse-dynamics model to predict action sequences
from pixels, combining an inverse model with a learned action prior.
LeFlow differs in three respects: 1) we operate in the latent space of a
\emph{frozen pretrained} world model rather than jointly learning representation
and planner; 2) instead of one-shot action prediction, we generate an explicit
multi-step \emph{latent trajectory} with a flow model and decode actions locally
at each step -- separating high-level planning from low-level
control~\citep{park2024hiql}; and 3) we sample $N$ diverse candidates and
rerank by world-model rollout, grounding proposals in the predictor's dynamics.
RC-aux~\citep{li2026rcaux} is a concurrent approach that improves latent
geometry via auxiliary reachability supervision; our planner is orthogonal and
could be applied on top of such representations.
Model-based offline RL~\citep{yu2020mopo,kidambi2020morel} is also related, but
typically learns task-specific policies or value functions from reward
supervision. LeFlow instead amortizes
goal-conditioned planning without a task-specific reward; combining these
directions is an interesting direction for future work.

\paragraph{Generative models for trajectory planning.}
Diffuser~\citep{janner2022diffuser} and Decision Diffuser~\citep{ajay2023decision}
diffuse over joint state-action trajectories; Diffusion
Policy~\citep{chi2023diffusion} generates action chunks for behavior cloning.
LeFlow instead generates \emph{latent state paths} inside a frozen world model
and decodes actions only afterwards via inverse dynamics.
Planning is inherently multimodal -- multiple valid trajectories connect the same
start and goal -- so a generative model is essential; deterministic regression
collapses to a mean path.
We use rectified flow~\citep{liu2023rectified,lipman2023flow} rather than
diffusion because its straight-line transport can be integrated in very few Euler
steps, keeping per-candidate online cost low when $N$ paths must be generated and
rolled out at each replanning step.

\section{Method}
\label{sec:method}

The central design question in LeFlow is \emph{what space to plan in}.
A na\"{i}ve generative planner generates action sequences directly, but
actions are the wrong abstraction for a reusable prior: they are
high-dimensional and highly multimodal (many kinematically distinct
trajectories realize the same sub-goal), and they are decoupled from the
world model's internal geometry -- the encoder has already organized
task-relevant structure into a compact latent space, yet an action-space
generator must re-derive this structure from scratch.

Latent state trajectories avoid both problems. The encoder collapses
action-level multiplicity: many action sequences that move $o_t$ to
$o_{t+1}$ share a single latent transition $z_t\to z_{t+1}$. Planning in
this space is 1) lower-dimensional and smoother, 2) already organized
around the predictor's dynamics, and 3) action-agnostic -- the planner
reasons about \emph{where the agent should be}, delegating \emph{how it
gets there} to a lightweight inverse dynamics decoder. This separation
decouples multi-step goal-directed reasoning from low-level control,
allowing each module to be trained and improved independently.

LeFlow realizes this design with three learned components on a
\emph{frozen} LeWM backbone: (1) a \emph{rectified-flow latent-path
planner} that generates goal-conditioned latent trajectories; (2) an
\emph{inverse dynamics decoder} that converts latent transitions into
executable action chunks; and (3) a \emph{rollout reranking} stage that
validates generative proposals against the frozen world model's dynamics.
We describe each in turn, followed by training and inference.

\subsection{Preliminary}
\label{sec:method:backbone}

LeWM~\citep{maes2026lewm} provides two components that LeFlow uses
unchanged. The \emph{encoder} $\mathrm{enc}_\theta$ maps a pixel
observation $o$ to a compact latent embedding $z = \mathrm{enc}_\theta(o)$.
The \emph{predictor} $\mathrm{pred}_\phi$ models latent dynamics
autoregressively,
\begin{equation}
  \hat z_{t+1} = \mathrm{pred}_\phi(\hat z_t, a_t),
\end{equation}
so given a start latent and an action sequence, LeWM rolls out a predicted
latent trajectory. LeWM's own planner solves the finite-horizon
optimal-control problem $a^\star_{1:H} = \arg\min_{a_{1:H}} \lVert \hat
z_H - z_g \rVert_2^2$ online at every replanning step using
CEM~\citep{rubinstein1999cem}. We keep the backbone frozen and replace
only this online search. LeWM serves three roles in LeFlow: a
representation backbone that supplies the latent geometry, a dynamics
prior whose structure we distill into the trajectory generator, and a
rollout-based feasibility verifier that grounds generative proposals in
realizable dynamics.

\subsection{Latent Flow planner}
\label{sec:method:flow}

Let $z_{\text{start}} = \mathrm{enc}_\theta(o_{\text{cur}})$ and
$z_{\text{goal}} = \mathrm{enc}_\theta(o_{\text{goal}})$ be the encoded
current and goal observations. A \emph{latent path} of horizon $H$ is a
sequence $z_{0:H} = (z_0, z_1, \dots, z_H)$ whose endpoints are clamped
by construction to $z_0 = z_{\text{start}}$ and $z_H = z_{\text{goal}}$.
The planner generates only the $H{-}1$ \emph{interior} latent states.
Fixing the endpoints is a deliberate design choice: it injects the
goal-conditioning directly into the geometric structure of the generated
path, rather than relying on the model to output a path that \emph{happens
to end} near the goal. The model's capacity is thus fully spent on how
the latent state should evolve between two known anchors -- the
\emph{qualitative shape} of the trajectory -- which is precisely where
task-level planning knowledge lives.

We model the interior path with a conditional rectified
flow~\citep{liu2023rectified,lipman2023flow}. The choice of generative
model family matters for online efficiency. Diffusion models denoise
through many small steps, so generating $N$ candidates at every replanning
step incurs a cost that scales with both $N$ and the number of denoising
iterations. Rectified flow instead learns a velocity field $v_\psi$ that
transports a Gaussian noise sample $u_0 \sim \mathcal{N}(0,I)$ to a data
sample $u_1$ along the \emph{straight} interpolant $u_\tau =
(1{-}\tau)u_0 + \tau u_1$ by matching the constant target velocity $u_1 -
u_0$. Because the learned trajectories in function space are approximately
linear, the ODE can be integrated accurately with very few Euler steps --
typically 16 in our setting -- regardless of $N$. This is well suited to
latent paths: the encoder produces a smooth, low-dimensional space where
straight-line interpolations between nearby embeddings are geometrically
meaningful, making the rectified-flow assumption a natural fit.
The training objective is
\begin{equation}
  \mathcal{L}_{\text{flow}}
  = \mathbb{E}_{\tau,\, u_0,\, u_1}
    \big\lVert v_\psi\!\left(u_\tau, \tau \mid z_{\text{start}},
    z_{\text{goal}}\right) - (u_1 - u_0) \big\rVert_2^2 ,
\end{equation}
where $u_1$ is the flattened interior of a ground-truth latent path from
the offline dataset. At inference, $N$ diverse interiors are sampled in
parallel by integrating the learned ODE from $N$ independent noise draws.

\subsection{Inverse dynamics decoder}
\label{sec:method:invdyn}

A latent path specifies \emph{where} the agent should be at each step but
says nothing about \emph{which actions get it there}. Rather than trying
to jointly generate actions alongside latent states -- which would
re-introduce the multimodality problem in the generative model -- we
delegate action recovery to a dedicated \emph{inverse dynamics decoder}
$g_\omega$ that operates locally on each latent transition:
\begin{equation}
  a_t = g_\omega\!\big([\, z_t,\; z_{t+1},\; z_{t+1} - z_t \,]\big).
\end{equation}
The decoder takes the current and next latent states together with their
\emph{displacement} $z_{t+1} - z_t$ as an explicit feature. The
displacement matters: it provides directional information about the
transition that is not separately recoverable from $z_t$ or $z_{t+1}$
alone, and empirically improves the conditioning of the regression
without adding parameters. The inverse dynamics problem is
much better posed in latent space than in observation space: because the
encoder has collapsed many distinct observations to the same latent, the
mapping from a latent transition to the corresponding action chunk is
far less ambiguous than a mapping from raw pixel pairs.

Critically, this separation means the planner and the decoder solve
genuinely different problems at different levels of abstraction. The flow
model reasons about multi-step goal-directed structure -- the \emph{shape}
of the latent trajectory -- while the decoder answers the purely local
question of which action realizes a given latent step. Neither module
needs to solve the other's problem, and both can be improved or replaced
independently.

The decoder is trained with a mean-squared error loss against the
dataset's normalized action chunks:
\begin{equation}
  \mathcal{L}_{\text{inv}}
  = \mathbb{E}\,\big\lVert
    g_\omega\!\left([z_t,\,z_{t+1},\,z_{t+1}-z_t]\right) - a_t
    \big\rVert_2^2 .
\end{equation}

\subsection{Rollout reranking for controllable manifold}
\label{sec:method:rerank}

\begin{figure}[tbp]
  \centering
  \includegraphics[width=\linewidth]{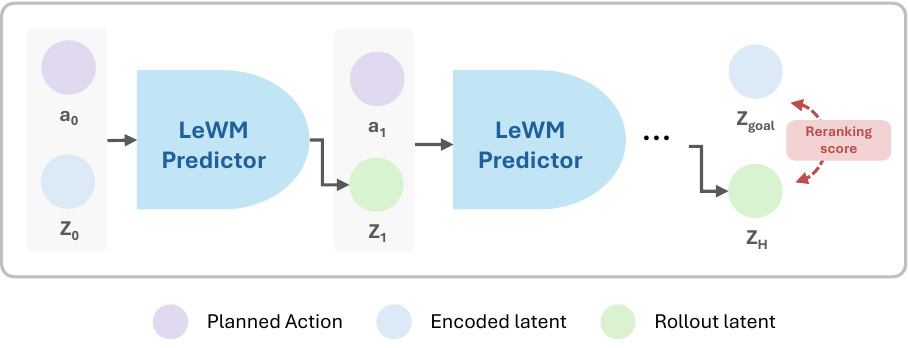}
  \caption{\textbf{Rollout reranking projects generative proposals onto the
  controllable latent manifold.}
  The flow model samples $N$ candidate latent paths (dashed) with endpoints
  clamped to $z_{\text{start}}$ and $z_{\text{goal}}$, but generated interiors
  may pass through latent states unreachable by any admissible action sequence.
  Each candidate is decoded into actions and rolled out autoregressively through
  the frozen LeWM predictor (solid), producing an \emph{actual} terminal latent
  $\hat{z}_H^{(i)}$. Candidates are ranked by rollout distance to the goal
  (Eq.~\ref{eq:rerank}); the best is executed. This step grounds amortized
  proposals in the predictor's dynamics without modifying the flow model.}
  \label{fig:reranking}
\end{figure}

Generative models do not respect the world model's dynamics by
construction: the flow model can interpolate through latent states that
no admissible action sequence actually reaches. Even a geometrically
reasonable path -- one that curves smoothly from $z_{\text{start}}$ to
$z_{\text{goal}}$ -- may pass through regions of latent space that the
predictor's learned dynamics cannot follow. We call the set of latent
trajectories that \emph{are} reachable by some action sequence the
\emph{controllable latent manifold}, and we use the frozen LeWM predictor
to project our generative proposals back onto it (Figure~\ref{fig:reranking}).

At inference we sample $N$ candidate paths from the flow model, decode
each into an action sequence, and roll the actions
\emph{autoregressively} through the frozen predictor to obtain the true
predicted terminal latent $\hat z_H^{(i)}$. Candidates are scored by
their rollout distance to the goal:
\begin{equation}
  \mathrm{score}^{(i)} = \big\lVert \hat z_H^{(i)} - z_{\text{goal}}
  \big\rVert_2^2 .
  \label{eq:rerank}
\end{equation}
Note that lower score indicates a better candidate -- the rollout landed
closer to the goal. We therefore select $\arg\min_i\,\mathrm{score}^{(i)}$
and execute its decoded action sequence. This reranking step is
inference-time and selection-only -- it does not alter the flow model's
distribution, only chooses among its samples.

\subsection{Training}
\label{sec:method:training}

LeWM is frozen throughout; LeFlow trains only the flow planner $v_\psi$
and the inverse dynamics decoder $g_\omega$, jointly, on offline
trajectory data. The total objective is
\begin{equation}
  \mathcal{L}
  = \mathcal{L}_{\text{flow}}
  + \mathcal{L}_{\text{inv}}
  + \lambda_{\text{cons}}\,\mathcal{L}_{\text{cons}} .
\end{equation}

\begin{table*}[tbp]
  \centering
  \caption{\textbf{Main benchmark results: success rate (\%) on four
  goal-conditioned pixel-control tasks.} Baseline numbers (goal-conditioned BC
  and offline RL -- GCBC, GCIVL, GCIQL; the JEPA world model PLDM; and the
  foundation-encoder world model DINO-WM) are as reported by
  \citet{maes2026lewm}; ``--'' marks benchmarks for which a baseline is not
  reported. LeFlow is the mean\,$\pm$\,std over \emph{five}
  independent evaluation runs (seeds $42$--$46$) under the LeWM
  codebase-default protocol ($H{=}5$, $50$ episodes). Best per column in \textbf{bold}.}
  \label{tab:main}
  \resizebox{0.8\linewidth}{!}{%
  \begin{tabular}{lcccc}
    \toprule
    Method & TwoRoom & PushT & Reacher & OGBench-Cube \\
    \midrule
    Random      & 0.0   & 2.0   & 10.0  & 48.0  \\
    GCBC~\citep{ghosh2019learnreach}  & 100.0 & 75.0  & --    & 84.0  \\
    GCIVL~\citep{park2025ogbench}     & 100.0 & 33.0  & --    & 56.0  \\
    GCIQL~\citep{kostrikov2021iql}    & 100.0 & 20.0  & --    & 64.0  \\
    PLDM~\citep{sobal2025pldm}        & 97.0  & 78.0  & 78.0  & 65.0  \\
    DINO-WM~\citep{zhou2024dinowm}    & 100.0 & 74.0  & 79.0  & 86.0  \\
    \midrule
    \textbf{\textit{LeWM-based Method}}\\
    CEM~\citep{rubinstein1999cem}
                & 82.0$\,\pm\,$2.0 & 89.3$\,\pm\,$6.4
                & 68.0$\,\pm\,$9.2 & 73.3$\,\pm\,$9.0 \\
    iCEM~\citep{pinneri2021icem}
                & 88.0$\,\pm\,$2.0 & 84.7$\,\pm\,$3.1
                & 67.3$\,\pm\,$11.4 & 76.0$\,\pm\,$7.2 \\
    MPPI~\citep{williams2017mppi}
                & 71.3$\,\pm\,$6.4 & 60.7$\,\pm\,$4.2
                & 42.7$\,\pm\,$4.6 & 49.3$\,\pm\,$9.5 \\
    \textbf{LeFlow} & \textbf{100.0$\,\pm\,$0.0} & \textbf{95.2$\,\pm\,$3.0}
                  & \textbf{86.8$\,\pm\,$4.8} & \textbf{100.0$\,\pm\,$0.0} \\
    \bottomrule
  \end{tabular}
  }
\end{table*}

\paragraph{Flow matching loss $\mathcal{L}_{\text{flow}}$.}
Trains the flow planner to match the distribution of latent-path
interiors from the offline dataset. This is the primary learning signal
for \emph{what trajectories look like} in the world model's latent space.

\paragraph{Inverse dynamics loss $\mathcal{L}_{\text{inv}}$.}
Trains the decoder to map latent transitions to action chunks. Because the
decoder is trained on \emph{dataset} transitions, it is naturally aligned
with the regions of latent space that the offline data covers -- the same
regions the flow model learns to generate in.

\paragraph{Consistency loss $\mathcal{L}_{\text{cons}}$.}
The two losses above do not guarantee that generated latent paths are
dynamically realizable: $\mathcal{L}_{\text{flow}}$ teaches the planner
to mimic the distribution of \emph{recorded} transitions, and
$\mathcal{L}_{\text{inv}}$ teaches the decoder to invert them, but neither
explicitly links the planner's output to the predictor's learned dynamics.
The consistency loss closes this loop. For each generated transition
$z_t \to z_{t+1}$, we decode the action $\hat a_t = g_\omega([z_t,
z_{t+1}, z_{t+1}-z_t])$, roll it one step through the frozen predictor
to obtain $\hat z_{t+1} = \mathrm{pred}_\phi(z_t, \hat a_t)$, and
penalize the gap:
\begin{equation}
  \mathcal{L}_{\text{cons}}
  = \mathbb{E}\,\big\lVert \hat z_{t+1} - z_{t+1} \big\rVert_2^2 .
\end{equation}
This one-step constraint \emph{distills the predictor's dynamics into the
planner at training time}: it steers the flow model's distribution toward
latent transitions that the predictor can actually follow, progressively
reducing the fraction of generated paths that reranking must discard.
The consistency loss and rollout reranking are therefore complementary
rather than redundant -- the former shapes the \emph{training distribution}
toward the controllable manifold, while the latter selects the best sample
at inference. A small weight ($\lambda_{\text{cons}} = 0.1$) is sufficient
because the flow model already learns from in-distribution transitions;
too large a weight would collapse the planner's distribution onto
the predictor's myopic one-step rollout, forfeiting the multi-step
planning structure that makes the flow model useful. We ablate this
weight in Section~\ref{sec:exp:consistency}.

\subsection{Inference}
\label{sec:method:inference}

Online planning reduces to a single batched computation: encode the
current and goal observations, sample $N$ latent paths from the flow model
with a small number of Euler integration steps, decode each into an action
sequence, rerank by frozen-LeWM rollout distance (Eq.~\ref{eq:rerank}),
and execute the best action chunk. Planning runs inside a receding-horizon
MPC loop following LeWM's action normalization and rollout conventions.
Crucially, there is no online optimization loop: the iterative
search of CEM -- hundreds of candidate evaluations over many refinement
rounds, restarted from scratch at each replanning step -- is entirely
replaced by a single forward pass through the flow model followed by one
batch of rollout evaluations. The planning computation is thus dominated
by the $N$ parallel rollouts of the frozen predictor, which are
embarrassingly parallelizable on a GPU, explaining the order-of-magnitude
speedup over CEM reported in Section~\ref{sec:exp:runtime}.

\begin{table*}[tbp]
  \centering
  \caption{\textbf{Planning efficiency: LeWM+CEM vs.\ LeFlow.} Both methods use
  the same frozen LeWM backbone and the same evaluation protocol ($H{=}5$, $50$
  episodes). Success rates are reported as in Table~\ref{tab:main} (LeFlow:
  mean over five runs, seeds $42$--$46$; LeWM+CEM: mean\,$\pm$\,std over three
  seeds under \texttt{stable\_worldmodel} defaults); eval time for CEM and LeFlow are both five-run mean. LeFlow exceeds CEM success on
  every benchmark while reducing end-to-end planning time by roughly an order
  of magnitude. Speedup is the ratio of CEM to mean LeFlow eval time.}
  \label{tab:runtime}
  \resizebox{0.8\linewidth}{!}{%
  \begin{tabular}{llccc}
    \toprule
    Benchmark & Method & Success (\%) $\uparrow$ & Eval Time (s) $\downarrow$ & Speedup $\uparrow$ \\
    \midrule
    \multirow{2}{*}{TwoRoom}
      & LeWM + CEM & 82.0$\,\pm\,$2.0 & 224.78 & 1.0$\times$ \\
      & \textbf{LeFlow} & \textbf{100.0$\,\pm\,$0.0} & \textbf{15.58} & \textbf{14.4$\times$} \\
    \midrule
    \multirow{2}{*}{PushT}
      & LeWM + CEM & 89.3$\,\pm\,$6.4 & 198.92 & 1.0$\times$ \\
      & \textbf{LeFlow} & \textbf{95.2$\,\pm\,$3.0} & \textbf{17.42} & \textbf{11.4$\times$} \\
    \midrule
    \multirow{2}{*}{Reacher}
      & LeWM + CEM & 68.0$\,\pm\,$9.2 & 326.01 & 1.0$\times$ \\
      & \textbf{LeFlow} & \textbf{86.8$\,\pm\,$4.8} & \textbf{27.67} & \textbf{11.8$\times$} \\
    \midrule
    \multirow{2}{*}{OGBench-Cube}
      & LeWM + CEM & 73.3$\,\pm\,$9.0 & 224.62 & 1.0$\times$ \\
      & \textbf{LeFlow} & \textbf{100.0$\,\pm\,$0.0} & \textbf{50.23} & \textbf{4.5$\times$} \\
    \bottomrule
  \end{tabular}
  }
\end{table*}

\section{Experiments}
\label{sec:experiments}

Our experiments test whether planning can be amortized without loss of
quality. Concretely: amortized latent planning can replace online CEM
optimization (1) at comparable or better success and (2) at substantially lower
planning cost; (3) planning in latent state space beats planning directly in
action space; and (4) rollout verification is necessary for dynamically feasible
plans. Throughout, the world model is frozen, so every result is attributable to
planning alone. Generalization to held-out episodes is evaluated in
Appendix~\ref{sec:appendix:heldout}.

\paragraph{Setup.}
We evaluate on the four goal-conditioned pixel-control benchmarks of
LeWM~\citep{maes2026lewm}: \textbf{TwoRoom} (2D navigation), \textbf{PushT} (2D
block manipulation), \textbf{Reacher} (2-joint reaching), and
\textbf{OGBench-Cube} (3D cube manipulation), all with continuous actions
(see Appendix~\ref{sec:appendix:benchmarks} for full descriptions). For
every benchmark we use a single frozen LeWM backbone, never fine-tuned. The LeFlow planner uses a 4-layer Transformer encoder as the rectified-flow
velocity model and a 3-layer MLP as the inverse dynamics decoder (full
architecture and training details in Appendix~\ref{sec:appendix:arch}).
It is trained for $10$ epochs with horizon $H{=}5$ and action block $5$.
Evaluation follows the LeWM codebase (MPC execution, LeWM action
normalization): the codebase-default $50$-episode protocol for the main and
runtime comparisons, $200$ episodes for ablations. Inference uses $N{=}64$
sampled latent paths and $16$ flow integration steps.
This controlled suite covers all official LeWM benchmarks but does not establish
scaling to substantially more complex environments.

\subsection{Main results}
\label{sec:exp:main}

Table~\ref{tab:main} reports LeFlow against the baselines collected by
\citet{maes2026lewm}: goal-conditioned behavioral cloning (GCBC) and offline RL
(GCIVL, GCIQL); the JEPA world models PLDM and LeWM; and the
foundation-encoder world model DINO-WM~\citep{zhou2024dinowm}, as well as the
LeWM-based online optimizers CEM, iCEM, and MPPI. LeFlow and LeWM+CEM are
reported as mean\,$\pm$\,std over repeated evaluations.
LeFlow reaches $100.0\pm0.0\%$ on TwoRoom, $95.2\pm3.0\%$
on PushT, $86.8\pm4.8\%$ on Reacher, and $100.0\pm0.0\%$ on OGBench-Cube with a
single planner design, attaining the highest mean success on every benchmark
and improving over the LeWM+CEM control by $+18.0$, $+5.9$, $+18.8$, and $+26.7$
points respectively. Amortized latent planning is thus a viable replacement for
iterative online action-space optimization.
Figure~\ref{fig:qual} shows representative successful rollouts across all four
benchmarks, illustrating the purposeful, goal-directed behavior produced by the
amortized planner without iterative online action-space optimization.

\subsection{Planning efficiency}
\label{sec:exp:runtime}

LeFlow's central practical claim is efficiency: CEM's online search is
amortized into a learned prior, so online planning becomes a fixed batched
computation. Table~\ref{tab:runtime} compares both
methods under an identical protocol with the same frozen backbone.

LeFlow is $14.4\times$ faster on TwoRoom, $11.4\times$ on PushT, $11.7\times$
on Reacher, and $4.5\times$ on OGBench-Cube, while exceeding LeWM+CEM success on
every benchmark. End-to-end time includes environment stepping and rendering,
shared by both methods, so the planner-side speedup is understated by these
numbers. This is our strongest result: trading offline planner training for an
order-of-magnitude reduction in online planning time.

\subsection{Planner design ablation}
\label{sec:exp:actionflow}

LeFlow makes two design choices over a naive generative planner: it plans in
\emph{latent state space} (rather than action space), and it models the latent
path \emph{generatively} (rather than by deterministic regression). We isolate
each with a dedicated ablation that changes only that one axis while keeping the
frozen LeWM backbone, $H{=}5$, and protocol fixed. The \emph{action-flow}
baseline generates normalized action-chunk sequences directly with a
rectified-flow model (reranked by the same rollout score), with no intermediate
latent states and no inverse dynamics decoder -- isolating the value of planning
in latent space. The \emph{deterministic latent-path} baseline replaces the
rectified-flow planner with a deterministic regressor that predicts a single
latent path from the start and goal embeddings -- isolating the value of
generative modeling; being deterministic, it produces one candidate and admits
no rollout reranking by construction. Because TwoRoom and OGBench-Cube are
saturated -- all variants achieve ${\geq}99.5\%$ success, leaving no headroom
for discriminative comparison -- we report results on the non-saturated
benchmarks PushT and Reacher.

\begin{table}[tbp]
  \centering
  \caption{\textbf{LeFlow vs.\ two planner ablations.} \emph{Action-Flow}
  generates action chunks directly (no latent states or inverse dynamics) and
  uses the same rollout reranking as LeFlow;
  \emph{Det.-Latent} replaces the rectified-flow planner with a deterministic
  regressor (single candidate, no reranking). $H{=}5$, $200$ episodes; $\Delta$
  is LeFlow minus the variant (positive favors LeFlow). Non-saturated benchmarks
  only. See Sec.~\ref{sec:exp:actionflow}.}
  \label{tab:actionflow}
  \resizebox{\columnwidth}{!}{%
  \begin{tabular}{l|c|cc|cc}
    \toprule
    & \textbf{LeFlow} & \multicolumn{2}{c|}{Action-Flow} & \multicolumn{2}{c}{Det.-Latent} \\
    Benchmark & (\%) & (\%) & $\Delta$ & (\%) & $\Delta$ \\
    \midrule
    PushT   & \textbf{96.5} & 93.5 & $+3.0$ & 95.5 & $+1.0$ \\
    Reacher & \textbf{87.5} & 83.0 & $+4.5$ & 83.0 & $+4.5$ \\
    \bottomrule
  \end{tabular}
  }
\end{table}

Table~\ref{tab:actionflow} shows that removing either design choice degrades
success, and LeFlow is best on both benchmarks. Planning in latent space
outperforms direct action generation by $+3.0$ on PushT and $+4.5$ on Reacher,
supporting our thesis that action sequences are low-level and multimodal, so
generating them directly is harder than generating the smoother latent state
path and decoding actions afterward. Generative modeling also helps: the
deterministic latent-path planner trails LeFlow by $+1.0$ on PushT and $+4.5$ on
Reacher, consistent with planning being inherently multimodal -- many valid
paths connect the same start and goal, and a deterministic regressor is pulled
toward their average rather than committing to a single realizable route. The
two effects are complementary: latent-space planning and generative modeling
each contribute, and LeFlow combines both.

\subsection{Rollout reranking ablation}
\label{sec:exp:rerank}

Table~\ref{tab:rerank} ablates rollout reranking against a no-rerank variant
that executes the first sampled candidate. As with the action-flow comparison,
we restrict the analysis to PushT and Reacher, the two non-saturated
benchmarks; on TwoRoom and OGBench-Cube both variants already exceed
$99.5\%$ success, so any measured difference would be noise rather than signal.
On the benchmarks where headroom exists, reranking is essential: it improves
Reacher by $+10.5$ points ($77.0\!\to\!87.5$) and PushT by $+2.5$ points.
This matches the design rationale: generated latent paths are not guaranteed
dynamically realizable, and scoring by the \emph{actual} frozen-LeWM rollout
projects planning back onto the controllable latent manifold.

\subsection{Consistency-loss ablation}
\label{sec:exp:consistency}

We ablate the consistency weight $\lambda_{\text{cons}}$, which encourages
generated paths to agree with the predictor's rollout of the decoded actions.
Table~\ref{tab:consistency} shows performance is stable across the tested
range; we use the default $\lambda_{\text{cons}}{=}0.1$ for all main models.
Notably, $\lambda_{\text{cons}}{=}0.0$ already performs competitively,
suggesting that the flow loss and rollout reranking together are sufficient for
most cases -- the consistency loss provides a marginal but consistent gain on
Reacher ($+2.0$ points), where the latent dynamics are more sensitive to
off-manifold transitions.
At $\lambda_{\text{cons}}{=}1.0$, performance slightly degrades on both
benchmarks, consistent with our design rationale: too strong a constraint
collapses the flow model's diversity toward the predictor's myopic one-step
rollout, reducing the benefit of sampling $N$ diverse candidates for reranking.

\begin{table}[tbp]
  \centering
  \caption{\textbf{Rollout reranking ablation.} ``Full'' scores candidates by
  frozen-LeWM rollout distance to the goal; ``No rerank'' executes the first
  sampled candidate. $H{=}5$, $200$ episodes; $\Delta$ is Full minus No-rerank.
  Non-saturated benchmarks only. See Sec.~\ref{sec:exp:rerank}.}
  \label{tab:rerank}
  \resizebox{\columnwidth}{!}{%
  \begin{tabular}{l|ccc}
    \toprule
    Benchmark & Full Rerank (\%) & No Rerank (\%) & $\Delta$ \\
    \midrule
    PushT        & 96.5  & 94.0  & $+2.5$  \\
    Reacher      & 87.5  & 77.0  & $+10.5$ \\
    \bottomrule
  \end{tabular}
  }
\end{table}

\begin{table}[tbp]
  \centering
  \caption{\textbf{Consistency-loss ablation.} Success rate (\%) as the
  consistency weight $\lambda_{\text{cons}}$ varies; $\lambda_{\text{cons}}{=}0.1$
  (main) is used by all main models. $H{=}5$, $200$ episodes.
  See Sec.~\ref{sec:exp:consistency}.}
  \label{tab:consistency}
  \resizebox{\columnwidth}{!}{%
  \begin{tabular}{lccc}
    \toprule
     & $\lambda_{\text{cons}}{=}0.0$ & $\lambda_{\text{cons}}{=}0.1$ (main) & $\lambda_{\text{cons}}{=}1.0$ \\
    \midrule
    PushT   & 96.5 & 96.5 & 95.0 \\
    Reacher & 85.5 & 87.5 & 84.5 \\
    \bottomrule
  \end{tabular}
  }
\end{table}

\begin{figure*}[t]
  \centering
  \includegraphics[width=\linewidth]{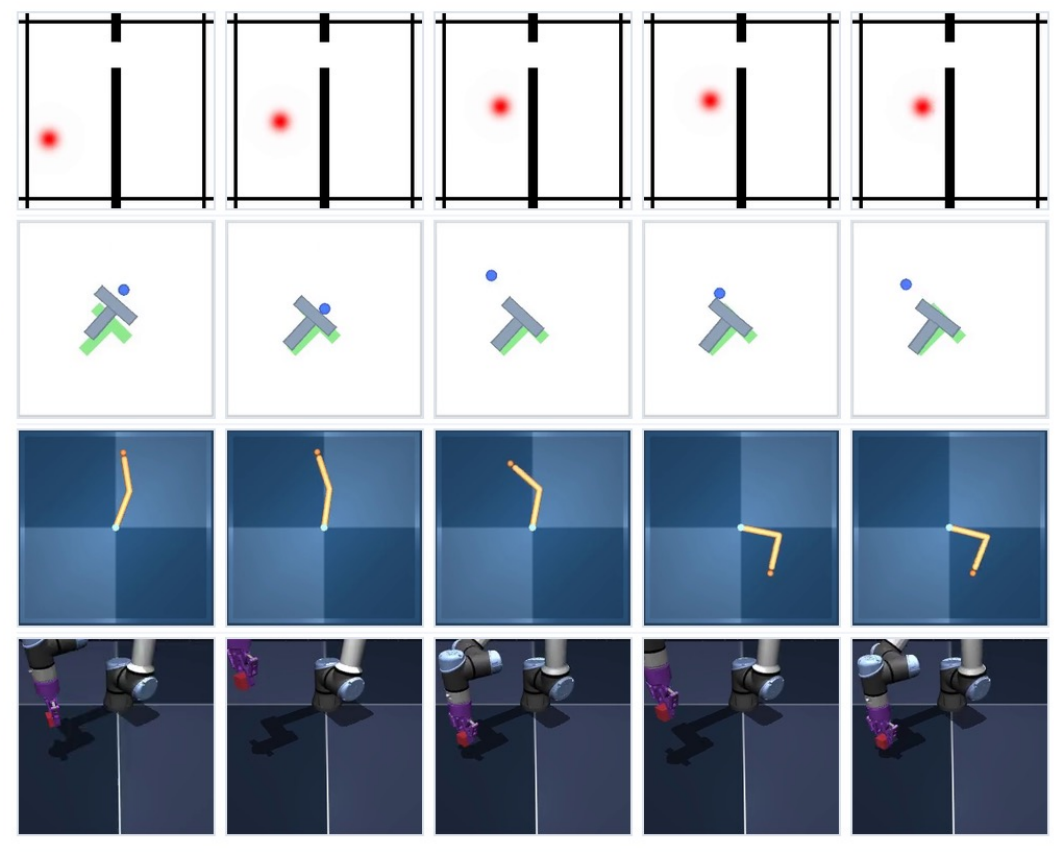}
  \caption{\textbf{LeFlow rollout visualizations.}
  Each row shows a single successful episode: start observation (left) through
  intermediate executed steps to goal achievement (right). Across navigation (TwoRoom), pushing (PushT), reaching (Reacher), and 3D
  manipulation (OGBench-Cube), LeFlow produces consistent,
  goal-directed behavior.}
  \label{fig:qual}
\end{figure*}

\subsection{Qualitative results}
\label{sec:exp:qual}

Figure~\ref{fig:qual} shows successful rollouts across all four benchmarks,
with each column representing a planning step and each cell showing the current
observation (top) alongside the goal (bottom).
In \textbf{TwoRoom}, the agent (red dot) begins in the upper room and must
navigate through the narrow doorway to reach a goal in the lower room; LeFlow
produces a smooth, door-aware trajectory that avoids the wall and converges
directly to the target position.
In \textbf{PushT}, the agent (blue dot) pushes the T-shaped block from an
initial misaligned pose to match the green target configuration, requiring
coordinated multi-step contact interactions; the planner correctly reasons
about both position and orientation of the block.
In \textbf{Reacher}, the two-joint arm progressively reconfigures from its
initial pose to match the goal joint configuration, producing smooth
coordinated joint motion without any oscillation or overshoot.
In \textbf{OGBench-Cube}, the robotic arm approaches, contacts, and
repositions the cube to match the goal state across the 3D scene.
Across all environments, the planner produces purposeful, goal-directed
behavior in a single forward pass, requiring no iterative action-space planning.

\section{Conclusion and Future Work}
\label{sec:conclusion}

In this work, we argued that planning on a latent world model need not be re-solved from
scratch. LeFlow \emph{amortizes} planning into a
reusable latent trajectory prior: on a frozen LeWM it generates goal-conditioned
latent paths, decodes actions via inverse dynamics, and verifies candidates by
frozen-model rollout. Across four
pixel-control benchmarks this replaces iterative action-space optimization with
one proposal-and-reranking pass, improving success while cutting planning time by roughly an order
of magnitude. However, LeFlow currently operates at a fixed short horizon. Longer horizons
increase the dimensionality of the generation problem and cause the frozen
predictor to accumulate rollout error over more steps, widening the gap between
generated paths and the controllable manifold. A natural direction is a
hierarchical decomposition that chains shorter-horizon LeFlow segments, or
coupling LeFlow with a stronger world model that reduces long-range rollout
error.


{
    \small
    \bibliographystyle{ieeenat_fullname}
    \bibliography{references}
}

\clearpage

\appendix

\section{Benchmark Descriptions}
\label{sec:appendix:benchmarks}

We evaluate on four goal-conditioned pixel-control benchmarks, all with
continuous action spaces. In every case the goal observation is a pixel image
drawn from the same episode as the start, sampled a fixed number of steps
ahead to ensure reachability. We follow the dataset and evaluation protocol
of \citet{maes2026lewm} exactly.

\paragraph{TwoRoom.}
A 2D navigation environment~\citep{sobal2025pldm} consisting of two rooms
separated by a wall with a single connecting door. The agent (a red dot) must
navigate from a random starting position in one room to a randomly sampled
target in the other room, necessarily passing through the doorway. The task
requires non-trivial multi-step planning: a straight-line path to the goal is
blocked by the wall, forcing the agent to first move toward the door and then
redirect toward the target. The offline dataset contains 10{,}000 episodes with
an average length of 92 steps, collected with a noisy heuristic policy. The
evaluation budget is 150 steps with the goal state sampled 100 steps into the
future. Action dimension: 2.

\paragraph{PushT.}
A 2D manipulation task in which an agent (a blue dot) must push a T-shaped
block to match a target configuration; contact is limited to pushing, with no
grasping. The task is challenging because the agent must reason about the
block's pose relative to the target and plan multi-step push sequences that
correct both position and orientation. We use the same 20{,}000 expert episodes
(average length 196 steps) as \citet{zhou2024dinowm}. The evaluation budget is
50 steps with the goal sampled 25 steps ahead. Action dimension: 2.

\paragraph{Reacher.}
A continuous control task from the DeepMind Control Suite in which a two-joint
planar robotic arm must reach a randomly placed target. The agent controls joint
torques; success requires coordinated multi-joint motion to place the
end-effector at the goal position. The task is visually sparse -- the scene
contains only the arm and a small target indicator -- making it a test of
dynamics modeling rather than visual complexity. The evaluation budget is 50
steps with the goal sampled 25 steps ahead. Action dimension: 2.

\paragraph{OGBench-Cube.}
A 3D manipulation environment from OGBench~\citep{park2025ogbench} in which a
robotic arm must pick up a cube and place it at a target position. Compared to
the 2D benchmarks, this task is significantly more visually complex: the scene
is rendered in 3D with realistic lighting and the cube's pose varies in all six
degrees of freedom. We use the single-cube variant with 10{,}000 episodes of
200 steps each, collected with the benchmark's default heuristic policy. The
evaluation budget is 50 steps with the goal sampled 25 steps ahead. Action
dimension: 4.

\section{Held-Out Generalization}
\label{sec:appendix:heldout}

To verify that the learned latent trajectory prior does not merely memorize
start--goal configurations seen during training, we evaluate LeFlow under an
\textbf{episode-level 80/20 held-out split}: the planner (flow model and inverse
dynamics decoder) is retrained using only the 80\% training partition of
episodes, and success rate is measured exclusively on start--goal queries drawn
from the held-out 20\% of episodes that were entirely excluded from planner
training. The frozen LeWM backbone is identical in both settings and is never
modified.

Table~\ref{tab:heldout} reports the results under the same 5-seed,
50-episode evaluation protocol used in the main comparison. Held-out success
rates are statistically indistinguishable from in-distribution performance on
every benchmark: TwoRoom $100.0\pm0.0\%$ (vs.\ $100.0\pm0.0\%$), PushT
$97.2\pm1.1\%$ (vs.\ $95.2\pm3.0\%$), Reacher $86.0\pm5.7\%$
(vs.\ $86.8\pm4.8\%$), and OGBench-Cube $100.0\pm0.0\%$ (vs.\
$100.0\pm0.0\%$). The negligible gap -- and the slight improvement on PushT --
indicates that the planner is not overfitting to specific episode trajectories.

This result is expected given the design: the flow model learns a distribution
over latent path \emph{shapes} in a frozen world model's embedding space, not a
lookup table over specific start--goal pixel pairs. The latent space organizes
observations by task-relevant geometry, so paths learned from one subset of
episodes provide useful inductive structure for novel start--goal pairs
elsewhere in the same latent space. The episode-level split (rather than a
clip-level split) provides a strict test because entire interaction sequences
are withheld, preventing any partial overlap between training and evaluation
contexts.

\begin{table}[h]
  \centering
  \caption{\textbf{Held-out generalization.} LeFlow retrained on 80\% of
  episodes and evaluated exclusively on the held-out 20\%. Results are
  mean\,$\pm$\,std over five seeds; in-distribution numbers (from
  Table~\ref{tab:main}) shown for reference. Differences are within noise on
  every benchmark.}
  \label{tab:heldout}
  \resizebox{\columnwidth}{!}{%
  \begin{tabular}{lcccc}
    \toprule
    & TwoRoom & PushT & Reacher & OGBench-Cube \\
    \midrule
    LeFlow (in-dist.)  & $100.0\pm0.0$ & $95.2\pm3.0$ & $86.8\pm4.8$ & $100.0\pm0.0$ \\
    LeFlow (held-out)  & $100.0\pm0.0$ & $97.2\pm1.1$ & $86.0\pm5.7$ & $100.0\pm0.0$ \\
    \bottomrule
  \end{tabular}
  }
\end{table}

\section{Architecture and Training Details}
\label{sec:appendix:arch}

\paragraph{Rectified-flow latent-path planner.}
The velocity model $v_\psi$ is a \textbf{4-layer Transformer encoder}
(pre-norm, GELU activations) with hidden dimension $d{=}512$, $8$ attention
heads, and feedforward width $4d{=}2048$.
Each noisy interior token $u_\tau^{(i)}$ is projected from latent space to
$\mathbb{R}^{512}$ via a learned linear layer and combined with a learned
positional embedding (supporting up to 19 interior steps, i.e.\
$H_{\max}{=}20$).
Conditioning on the flow time $\tau$ uses a sinusoidal embedding of dimension
64 passed through two linear layers with SiLU activation; the resulting
time vector is summed with the start and goal projections and broadcast
across all tokens.
The output head is a \texttt{LayerNorm} followed by a linear projection back
to latent dimension.


\paragraph{Inverse dynamics decoder.}
The decoder $g_\omega$ is a \textbf{3-layer MLP} with hidden dimension 512,
\texttt{LayerNorm}$+$\texttt{GELU} after each hidden layer, and no dropout.
Its input is the concatenation $[z_t,\,z_{t+1},\,z_{t+1}{-}z_t]$ of dimension
$3{\times}\text{latent\_dim}$.



\paragraph{Training hyperparameters.}
Both modules are trained jointly with AdamW ($\text{lr}{=}10^{-4}$,
$\text{weight\_decay}{=}10^{-4}$) for 10 epochs with batch size 128,
cosine-annealing learning rate schedule, and gradient clipping at norm 1.0.
Loss weights: $\lambda_{\text{flow}}{=}1.0$,
$\lambda_{\text{inv}}{=}1.0$, $\lambda_{\text{cons}}{=}0.1$.
Inference uses $N{=}64$ sampled paths and 16 Euler integration steps.

\section{Horizon-scaling Results}

\begin{table}[h]
  \centering
  \caption{\textbf{PushT horizon scaling.} Each planner is trained separately
  at its reported horizon. All settings use 50 evaluation cases sampled
  with seed 42, a 100-step budget, and an action block of five.}
  \label{tab:long-horizon}
  \resizebox{0.5\columnwidth}{!}{%
  \begin{tabular}{ccc}
    \toprule
    H & Success (\%) & Time (s) \\
    \midrule
    5  & 94.0 & 28.00 \\
    10 & 32.0 & 30.57 \\
    20 & 6.0  & 34.73 \\
    \bottomrule
  \end{tabular}
  }
\end{table}

The main experiments use a planner horizon of $H{=}5$, with each latent
transition decoded into an action block of five environment steps. To examine
scaling beyond this setting, we train otherwise identical PushT planners at
$H\in\{5,10,20\}$. Each planner uses the same frozen LeWM checkpoint and
35,500 training updates. At evaluation, we keep the action block at five. All
settings use 50 evaluation cases sampled with seed 42, a 100-step execution
budget, a receding horizon of five latent action blocks, 64 proposals, 16 flow
steps, and rollout-goal reranking.

Table~\ref{tab:long-horizon} reports success and planning runtime. Success decreases from $94\%$ at $H{=}5$ to $32\%$ at $H{=}10$ and
$6\%$ at $H{=}20$, while runtime increases from 28.00 to 34.73
seconds. The degradation is expected due to two factors. First, longer paths
are intrinsically harder for the current model to learn: the flow model must
coordinate more latent transitions, the inverse model must decode more action
blocks, and prediction errors can accumulate over longer rollouts. Second,
PushT and other official LeWM benchmarks were designed primarily for short-term
control rather than for measuring long-horizon scaling. Increasing the horizon
also produces longer and less stable training trajectories and moves the
evaluation away from the benchmark's standard short-term protocol. The observed decrease should
therefore be understood as a combination of limitations of the current method
and limitations of the benchmark for studying long-horizon control.
Nonetheless, we emphasize that our goal in this work is not to extend LeWM to
long-horizon planning tasks, but to determine whether its online trajectory
optimization can be amortized through learned latent-path generation. Exploring
methods such as hierarchical planning and benchmarks that support long-term
planning evaluation are promising future directions.


\begin{table}[ht]
  \centering
  \caption{\textbf{LeFlow on a frozen DINO-WM backbone (PushT).} All methods
  use the same 50 evaluation cases. LeFlow rows use the same frozen spatial
  encoder and visual-plus-proprio representation. Time is end-to-end job time
  for all 50 cases; state distance is DINO-WM's final-state metric.}
  \label{tab:dino-scaling}
  \resizebox{\columnwidth}{!}{%
  \begin{tabular}{lcccc}
    \toprule
    Method (data) & Proposals & SR (\%) & Distance $\downarrow$ & Time (m:s) \\
    \midrule
    LeFlow (1k)  & 64 & 38.0 & 38.60 & 0:43 \\
    LeFlow (5k)  & 64 & 60.0 & 23.88 & 0:43 \\
    LeFlow (10k) & 64 & 68.0 & 26.53 & 0:34 \\
    LeFlow (15k) & 64 & 78.0 & 19.40 & 0:34 \\
    \textbf{LeFlow (15k)} & \textbf{512} & \textbf{86.0} &
      \textbf{15.55} & \textbf{2:32} \\
    \midrule
    DINO-WM+CEM  & $300{\times}30$ & 84.0 & 25.41 & 37:09 \\
    \bottomrule
  \end{tabular}
  }
\end{table}

\section{Generalization to DINO-WM}
\label{sec:appendix:dino-wm}

The main experiments deliberately hold LeWM fixed to isolate the effect of the
planner. To test whether the planner design depends specifically on LeWM's
latent representation, we additionally apply LeFlow to the official pretrained
DINO-WM~\citep{zhou2024dinowm} checkpoint on PushT. DINO-WM represents an image
with a $256\times384$ grid of DINOv2 patch features rather than the single
compact vector expected by our trajectory model. We therefore train a spatial
autoencoder that compresses the ordered patch grid into a 512-dimensional
visual vector. Its encoder consists of learned positional embeddings, a
summary token, and a two-layer Transformer; its training-only decoder
reconstructs the frozen patch features using MSE and cosine losses and is
discarded afterwards. The spatial autoencoder is trained on 124{,}323 images
from the first 1{,}000 PushT trajectories and remains fixed in all experiments
below.

The planner representation concatenates this visual vector with DINO-WM's
10-dimensional proprioceptive embedding,
\begin{equation}
  z_t = [z_t^{\mathrm{visual}};z_t^{\mathrm{proprio}}]
  \in \mathbb{R}^{522}.
\end{equation}
The proprioceptive input contains the agent position and velocity; the pushed
object's pose and orientation must still be inferred visually. LeFlow generates
an $H{=}5$ path in this compact space, and the inverse-dynamics model decodes
each transition into five low-level actions to match DINO-WM's frame skip. At
inference, these actions are passed to the original frozen DINO-WM
transition model, initialized from its full patch-token and proprioceptive
state, for candidate rollout and reranking.

We retain the default LeFlow architecture and optimization wherever possible:
a four-layer, width-512 flow Transformer, a three-layer, width-512 inverse
dynamics MLP, AdamW with learning rate and weight decay $10^{-4}$, 10 training
epochs, and 16 Euler steps. Evaluation follows DINO-WM's official
\texttt{PlanWorkspace}, dataset-goal construction, PushT rollout, and success
metric on the same 50 cases for every model. The 5k, 10k,
and 15k trajectory subsets are sampled once with fixed seed. Only the LeFlow
planner and inverse model are scaled; the DINO-WM checkpoint and spatial
encoder remain frozen. The matched baseline uses DINO-WM's official CEM
configuration: 300 samples, 30 elites, variance scale 1, and 30 optimization
iterations.

Table~\ref{tab:dino-scaling} shows monotonic success gains as planner-training
data increase: $38\%$ with 1k trajectories, $60\%$ with 5k, $68\%$ with 10k,
and $78\%$ with 15k, using 64 proposals throughout. Increasing only the
inference proposal count of the same 15k model to 512 raises success to
$86\%$, surpassing matched DINO-WM+CEM at $84\%$, while reducing mean state
distance from 25.41 to 15.55. LeFlow reranks its 512 candidates in one pass;
CEM performs 30 refinement iterations over 300 candidates, or 9{,}000
candidate transition rollouts per case. LeFlow therefore obtains the higher
success rate with $17.6\times$ fewer candidate rollouts. The measured end-to-end time for all 50 cases is 2:32 for LeFlow and 37:09 for CEM, a $14.7\times$ reduction.

Together, these results show that the latent-path proposal, local action
decoding, and frozen-world-model verification decomposition transfers beyond
LeWM and retains its main advantage over iterative optimization. Nonetheless, the experiments are limited to PushT and its spatial bottleneck is trained on only 1k
trajectories, so we view it as evidence of transfer to a second backbone rather
than universal backbone independence.

\end{document}